\documentclass[sigconf]{acmart}

\usepackage{booktabs} 
\usepackage{bm}
\usepackage{mathtools}
\usepackage{subcaption}

\usepackage{algorithm}
\usepackage{algorithmic}
\usepackage{color}

\setcopyright{rightsretained}

\acmDOI{10.1145/nnnnnnn.nnnnnnn}

\acmISBN{978-x-xxxx-xxxx-x/YY/MM}

\acmConference[GECCO '18]{the Genetic and Evolutionary Computation
Conference 2018}{July 15--19, 2018}{Kyoto, Japan}
\acmYear{2018}
\copyrightyear{2018}

\acmArticle{4}
\acmPrice{15.00}

\begin{document}
\title{Evolutionary Architecture Search For Deep Multitask Networks}

\author{Jason Liang, Elliot Meyerson, and Risto Miikkulainen}
\affiliation{%
  \institution{Sentient Technologies, Inc. and}
  \institution{The University of Texas at Austin}
}

\begin{abstract}
  Multitask learning, i.e. learning several tasks at once with the same neural network, can improve performance in each of the tasks. Designing deep neural network architectures for multitask learning is a challenge: There are many ways to tie the tasks together, and the design choices matter. The size and complexity of this problem exceeds human design ability, making it a compelling domain for evolutionary optimization. Using the existing state of the art soft ordering architecture as the starting point, methods for evolving the modules of this architecture and for evolving the overall topology or routing between modules are evaluated in this paper. A synergetic approach of evolving custom routings with evolved, shared modules for each task is found to be very powerful, significantly improving the state of the art in the Omniglot multitask, multialphabet character recognition domain. This result demonstrates how evolution can be instrumental in advancing deep neural network and complex system design in general.
\end{abstract}

%
%

\ccsdesc[500]{Computing methodologies~Neural networks}
\ccsdesc[300]{Computing methodologies~Distributed artificial intelligence}
\ccsdesc[100]{Computing methodologies~Computer vision}

\keywords{Neural Networks/Deep Learning, Artificial Intelligence}

\maketitle

\section{Introduction}

In multitask learning (MTL) , a neural network is trained simultaneously to perform several different tasks at once \cite{Caruana:1998}. For instance, given an image as input, it can recognize the objects in it, identify the type of scene, and generate a verbal caption for it. Typically the early parts of the network are shared between tasks, and the later parts, leading to the different tasks, are separate \cite{Caruana:1998, Collobert:2008, Dong:2015, Lu:2016, Ranjan:2016}.  The network is trained with gradient descent in all these tasks, and therefore the requirements of all tasks are combined in the shared parts of the network. The embeddings thus reflect the requirements of all tasks, making them more robust and general. Performance of a multitask network in each task can therefore exceed the performance of a network trained in only
a single task.

Much of the research in deep learning in recent years has focused on coming up with better architectures, and MTL is no exception. As a matter of fact, architecture plays possibly an even larger role in MTL because there are many ways to tie the multiple tasks together. The best network architectures are large and complex, and have become very hard for human designers to
optimize \cite{szegedy2015going, szegedy:cvpr16, zoph:arxiv16, jaderberg2017population}

This paper develops an automated, flexible approach for evolving architectures, i.e. hyperparameters, modules, and module routing topologies, of deep multitask networks. A recent deep MTL architecture called soft ordering \cite{Meyerson:2018} is used as a starting point, in which a different soft sequence of modules is learned for each task. This paper extends this architecture in several ways. First, this paper proposes a novel algorithm for evolving \emph{task specific routings} that create an unique routing between modules for each task. Second, \emph{more general modules} with the same soft ordering architecture are evolved.
Third, the general modules are evolved together with a \emph{blueprint}, a shared routing for all tasks, that improves upon the soft ordering architecture. Fourth, as a capstone architecture, the task specific routing are evolved together with the general modules.
These four approaches are evaluated in the Omniglot task \cite{Lake:2015} of learning to recognize characters from many different alphabets. A series of results confirms the intuition well: As a baseline, soft ordering performs significantly better in each task than single-task training (67\% vs.\ 61\% accuracy). Evolution of modules and topologies improves significantly upon soft ordering. Coevolution of modules and topologies together improves even more, and the capstone architecture turns out to be the best (at 88\%).

The results thus demonstrate three general points: evolutionary architecture search can make a large difference in performance of deep learning networks; MTL can improve performance of deep learning tasks; and putting these together results in a particularly powerful approach.  In the future it can be applied to various problems in vision, language, and control, and in particular to domains with multimodal inputs and outputs.
The rest of the paper is organized as follows: In Section~\ref{sec:background}, previous work on deep MTL and neural architecture search are summarized. In Section~\ref{sec:algorithms}, the key contribution of this paper, novel evolutionary algorithms for architecture search of multitask networks are described. Finally, in Section~\ref{sec:experiments} and Section~\ref{sec:discussion} experimental results on the Omniglot domain are presented and analyzed.

\section{Background and Related Work}
\label{sec:background}

Before introducing methods for combining them in Section~\ref{sec:algorithms}, this section reviews deep MTL and neural architecture search.

\begin{figure}
  \begin{center}
    \includegraphics[width=\linewidth, height=1.1in]{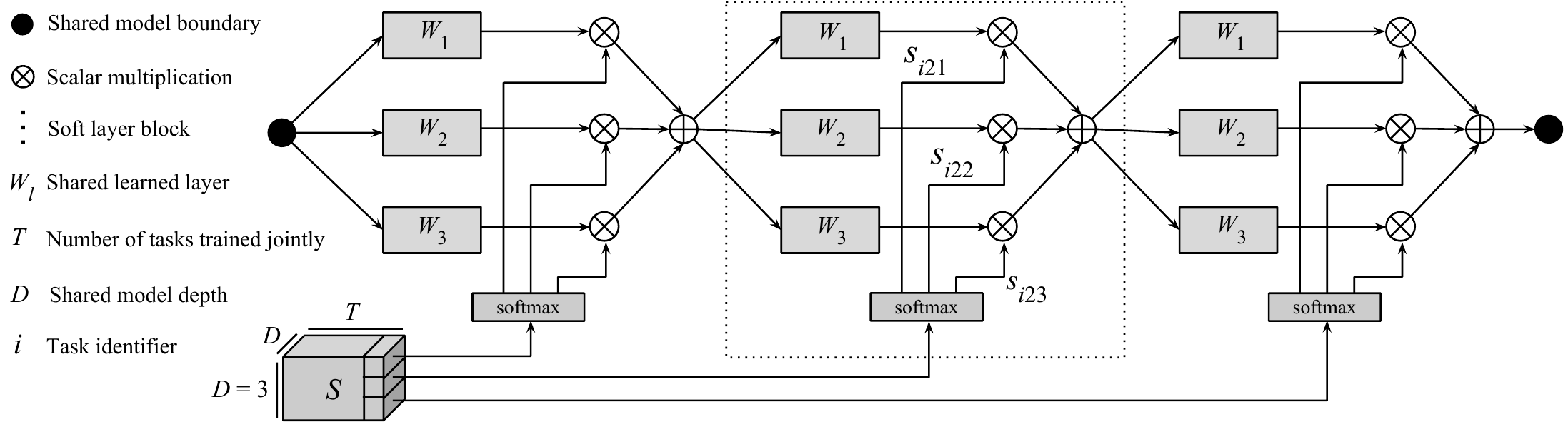}
    \caption{\label{fig:soft_order} Example soft ordering network with three shared layers. Soft ordering learns how to use the same layers in different locations by learning a tensor $S$ of task-specific scaling parameters. $S$ is learned jointly with the $W_d$, to allow flexible sharing across tasks and depths. This architecture enables the learning of layers that are used in different ways at different depths for different tasks.}
  \end{center}
\end{figure}

\subsection{Deep Multitask Learning}

MTL \cite{Caruana:1998} exploits relationships across problems to increase overall performance.
The underlying idea is that if multiple tasks are related, the optimal models for those tasks will be related as well.
In the convex optimization setting, this idea has been implemented via various regularization penalties on shared parameter matrices \cite{Argyriou:2008, Evgeniou:2004, Kang:2011, Kumar:2012}.
Evolutionary methods have also had success in MTL, especially in sequential decision-making domains \cite{Huizinga:2016, Kelly:2017, Jaskowski:2008, Schrum:2016, Snel:2010}.

Deep MTL has extended these ideas to domains where deep learning thrives, including vision \cite{Bilen:2017, Kaiser:2017, Lu:2016, Misra:2016, Ranjan:2016, Rebuffi:2017, Yang:2017, Zhang:2014}, speech \cite{Huang:2013, Huang:2015, Kaiser:2017, Seltzer:2013, Wu:2015}, natural language processing \cite{Collobert:2008, Dong:2015, Hashimoto:2016, Kaiser:2017, Liu:2015, Luong:2016, Zhang:2016}, and reinforcement learning \cite{Devin:2016, Jaderberg:2016, Teh:2017}.
The key design decision in constructing a deep multitask network is deciding how parameters such as convolutional kernels or weight matrices are shared across tasks.
Designing a deep neural network for a single task is already a high-dimensional open-ended optimization problem; having to design a network for multiple tasks \emph{and} deciding how these networks share parameters grows this search space combinatorially.
Most existing approaches draw from the deep learning perspective that each task has an underlying feature hierarchy, and tasks are related through an \emph{a priori} alignment of their respective hierarchies.
These methods have been reviewed in more detail in previous work \cite{Meyerson:2018, Ruder:2017}.
Another existing approach adapts network structure by learning task hierarchies, though it still assumes this strong hierarchical feature alignment \cite{Lu:2016}.

Soft ordering is a recent approach that avoids such an alignment by allowing shared layers to be used across different depths \cite{Meyerson:2018}.
Through backpropagation, the joint model learns how to use each shared (potentially nonlinear) layer $W_d$ at each depth $d$ for the $t$-th task.
This idea is implemented by learning a distinct scalar $s_{tdl}$ for each such location, which then multiplies the layer's output.
The final output at depth $d$ for the task is then the sum of these weighted outputs across layers, i.e., a \emph{soft merge}.
More generally, a soft merge is a learnable function given by
\begin{equation}
\label{eq:soft_merge}
\text{softmerge}(\text{in}_1, \ldots, \text{in}_M) = \sum_{m=1..M} s_{m}\text{in}_m, \text{ with } \sum_{m=1..M} s_{m} = 1 \, ,
\end{equation}
where the $\text{in}_m$ are a list of incoming tensors, $s_m$ are scalars trained simultaneously with internal layer weights via backpropagation, and the constraint that all $s_m$ sum to 1 is enforced via a softmax function.
Figure~\ref{fig:soft_order} shows an example soft ordering network.

More formally, given shared layers $W_1, W_2, \ldots, W_D$, the soft ordering model $\bm{y}_t = f(\bm{x}_t)$ for the $t$-th task $\{(\bm{y}_{ti}, \bm{x}_{ti})\}_{i=1}^N$ is given by $\bm{y}_t = \mathcal{D}_t(\bm{y}^D_t)$, where $\bm{y}^0_t = \mathcal{E}_t(\bm{x}_t)$ and
\begin{align}
	\bm{y}^{d}_t &= \text{softmerge}\big(W_1(\bm{y}^{d-1}_t), \ldots , W_D(\bm{y}^{d-1}_t)\big) \ \ \ \forall \ d \in 1..D\, , \label{eq:soft_order}
\end{align}
where $\mathcal{E}_t$ is a task-specific encoder mapping the task input to the input of the shared layers, $\mathcal{D}_t$ is a task-specific decoder mapping the output of the shared layers to an output layer, e.g., classification.

Although soft ordering allows flexible sharing across depths, layers are still only applied in a fixed grid-like topology, which biases and restricts the type of sharing that can be learned.
This paper generalizes soft ordering layers to more general modules, and introduces evolutionary approaches to both design these modules and to discover how to assemble these modules into appropriate topologies for multitask learning.
To our knowledge, this is the first paper that takes an evolutionary approach to deep MTL.

\begin{figure}[t]
  \begin{center}
    \includegraphics[width=\linewidth]{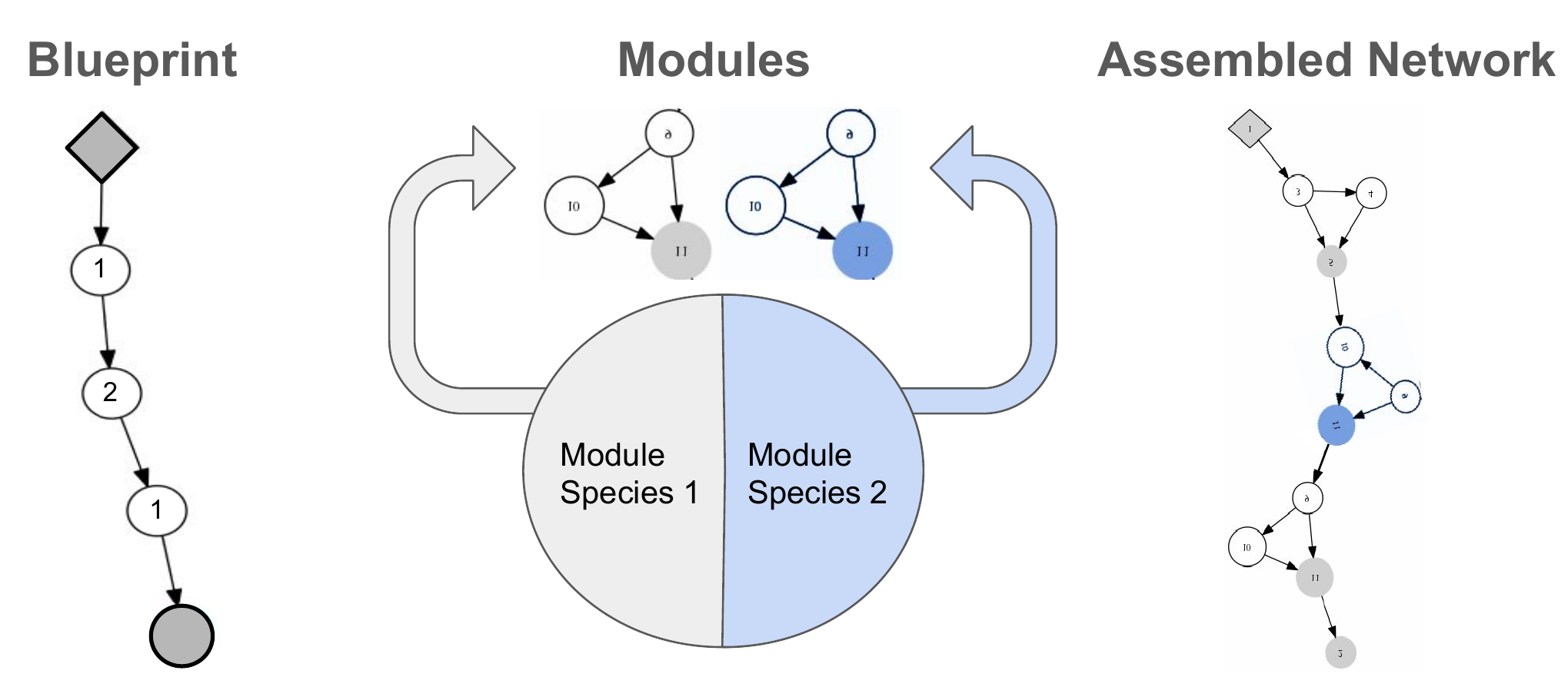}
    \caption{Assembling networks for fitness evaluation in CoDeepNEAT. Modules from species specified in the blueprint are inserted into locations specified in the blueprint, forming a full network. This approach allows evolving repetitive and deep structures seen in many successful DNNs.}
    \label{fg:assembling}
  \end{center}
\end{figure}

\subsection{Architecture and Hyperparameter Search}
\label{sc:nas}

As deep learning tasks and benchmarks become increasing complex, finding the right architecture becomes more important. In fact, the performance of many state of the art networks \cite{he:arxiv16,szegedy2015going,ng:arxiv15,szegedy:cvpr16} depend mostly on novel and interesting architectural innovations. Unfortunately, discovering useful hyperparameters and architectures by hand is tedious and difficult; as a result, much research focuses on developing automated methods for doing it. Some promising methods for hyperparameter search include deep Bayesian optimization \cite{snoek:arxiv15} and CMA-ES \cite{loshchilov:arxiv16}. One unique approach uses reinforcement learning to develop an LSTM policy for generating appropriate network topologies and hyperparameters for a given task \cite{zoph:arxiv16}.

One particular promising area of research is the use of evolutionary algorithms (EAs) for performing architecture search. Evolutionary methods are well suited for this kind of problems because they can be readily applied with no gradient information. Some of these approaches use a modified version of NEAT \cite{miikkulainen2017evolving,real2017large}, an EA for neuron-level neuroevolution \cite{stanley:ec02}, for searching network topologies. Others rely on genetic programming \cite{suganuma2017genetic} or hierarchical evolution \cite{jaderberg2017population}. Along these lines, CoDeepNEAT \cite{miikkulainen2017evolving} combines the power of NEAT's neural topology search with hierarchical evolution to efficiently discover architectures within large search spaces. Networks evolved using CoDeepNEAT have achieved good results in image classification and image captioning domains, outperforming popular hand-designed architectures. Consequently, CoDeepNEAT is one of the methods used in this paper for optimizing the topologies of the deep multitask networks. This paper extends CoDeepNEAT to coevolve modules and blueprints for deep multitask networks and also combines it with a novel and powerful EA for evolving task-specific architectures with shared modules.

CoDeepNEAT begins by initializing two populations, one of modules and one of blueprints, with minimal complexity. The blueprints and modules each contain at least one species and are evolved/complexified separately with a modified version of NEAT. An individual in the blueprint population is a directed acyclic graph (DAG) where each node contains a pointer to a particular module species. An individual in the module population is a DAG where each node represents a particular DNN layer and its corresponding hyperparameters (number of neurons, activation function, etc). As shown in Figure~\ref{fg:assembling}, the modules are inserted into the blueprints to create a temporary population of assembled networks. Each individual in this population is then evaluated by training it on a supervised learning task, and assigning its performance as fitness. The fitnesses of the individuals (networks) are attributed back to blueprints and modules as the average fitness of all the assembled networks containing that blueprint or module. One of the advantages of CoDeepNEAT is that it is capable of discovering modular, repetitive structures seen in state of the art networks such as Googlenet and Resnet \cite{szegedy2015going,szegedy:cvpr16,he:arxiv16}.

\begin{figure}[t]
  \begin{center}
    \includegraphics[width=\linewidth]{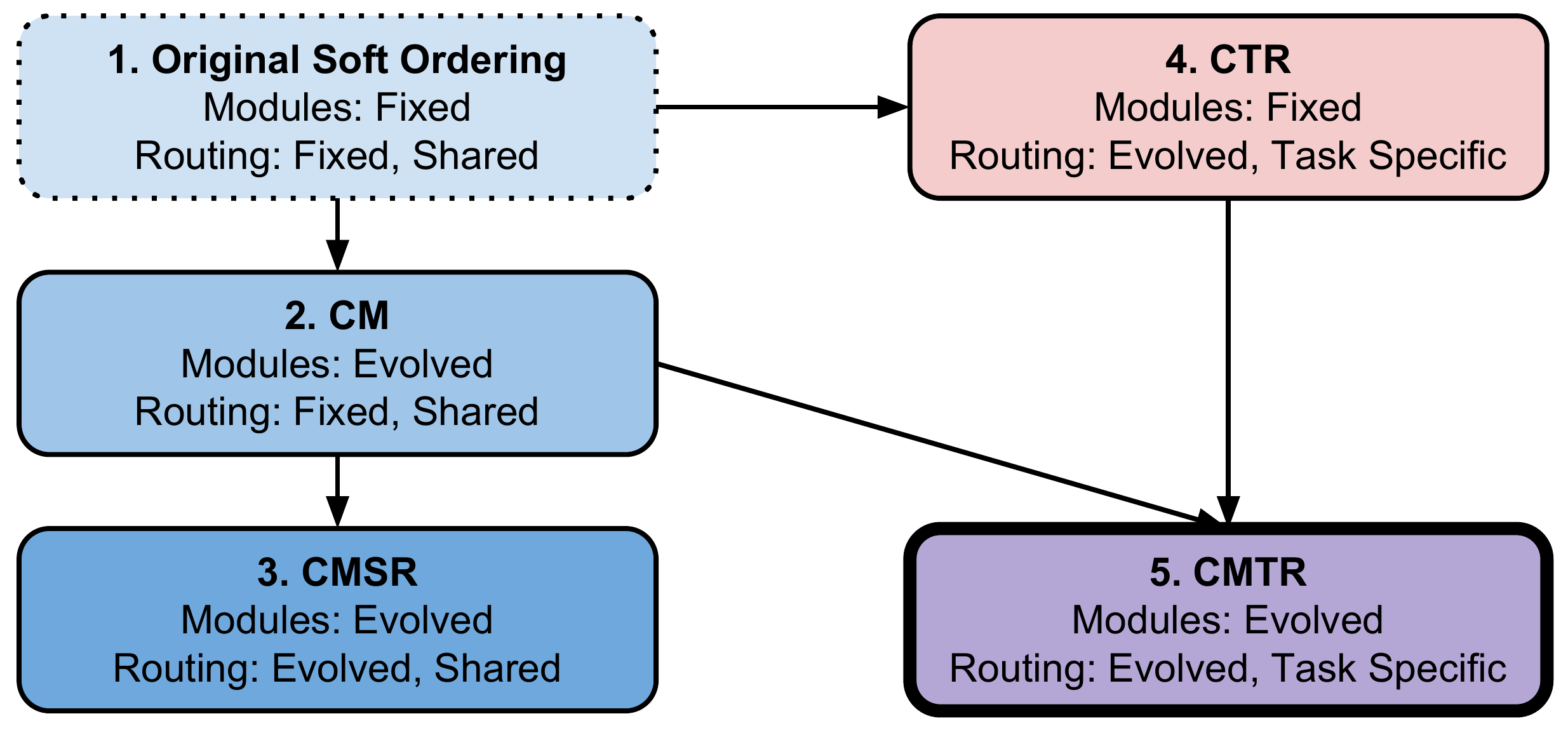}\\[-2ex]
    \caption{The relationships of the five methods tested in this paper. The soft ordering method \cite{Meyerson:2018} is used as the starting point, extending it with CoDeepNEAT on the left and task-specific routing on the right. The CMTR on bottom right combines these two main ideas and performs the best.}
    \label{fg:overview}
  \end{center}
\end{figure}

\begin{figure*}
\scriptsize
\begin{minipage}{1.7in}
\begin{algorithm}[H]
\caption{CM (Section~\ref{subsec:cm})}
\flushleft
\textbf{Given} fixed blueprint\\
\textbf{Initialize} module population\\
\textbf{Each} generation:\\
    \hspace{10pt} \textbf{Assemble} MTL networks with modules\\
    \hspace{10pt} \textbf{Randomly initialize all weights}\\
    \hspace{10pt} \textbf{Train} each MTL network with backprop\\
    \hspace{10pt} \textbf{Assign fitnesses} to modules\\
    \hspace{10pt} \textbf{Update} module populations
\end{algorithm}
\end{minipage}
\begin{minipage}{1.8in}
\begin{algorithm}[H]
\caption{CMSR (Sec.~\ref{subsec:cmsr})}
\flushleft
\textbf{Initialize} blueprint/module populations\\
\textbf{Each} generation:\\
    \hspace{10pt} \textbf{Assemble} MTL networks with\\
    \hspace{10pt} blueprints/modules\\
    \hspace{10pt} \textbf{Randomly initialize all weights}\\
    \hspace{10pt} \textbf{Train} each MTL network with backprop\\
    \hspace{10pt} \textbf{Assign fitnesses} to modules and blueprints\\
    \hspace{10pt} \textbf{Update} blueprint/module populations
\end{algorithm}
\end{minipage}
\begin{minipage}{1.7in}
\begin{algorithm}[H]
\caption{CTR (Sec.~\ref{subsec:ctr})}
\flushleft
\textbf{Given} set of modules\\
\textbf{Initialize} topology population for each task\\
\textbf{Randomly initialize all weights}\\
\textbf{Each} meta-iteration:\\
    \hspace{10pt} \textbf{Assemble} networks\\
    \hspace{10pt} Jointly \textbf{train} all networks with backprop\\
    \hspace{10pt} \textbf{Assign fitnesses} to topologies\\
    \hspace{10pt} \textbf{Update} topology populations
\end{algorithm}
\end{minipage}
\begin{minipage}{1.6in}
\begin{algorithm}[H]
\caption{CMTR (Sec.~\ref{subsec:cmtr})}
\flushleft
\textbf{Initialize} module population\\
\textbf{Each} generation:\\
    \hspace{10pt} \textbf{Assemble} sets of modules\\
    \hspace{10pt} \textbf{Train} sets of modules with CTR\\
    \hspace{10pt} \textbf{Assign fitnesses} to modules\\
    \hspace{10pt} \textbf{Update} module populations
\vspace{14pt}
\end{algorithm}
\end{minipage}
\caption{\label{fig:algs} High-level algorithm outlines of the four architecture evolution algorithms introduced in this paper, illustrating how they are related.
In particular, the algorithms differ in which components are evolved, and how they are trained.
}
\end{figure*}

\section{Algorithms for Deep MTL Evolution}
\label{sec:algorithms}

Figure~\ref{fg:overview} provides an overview of the methods tested in this paper in multitask learning. The foundation is (1) the original soft ordering, which uses a fixed architecture for the modules and a fixed routing (i.e.\ network topology) that is shared among all tasks.  This architecture is then extended in two ways with CoDeepNEAT: (2) by coevolving the module architectures (CM), and (3) by coevolving both the module architectures and a single shared routing for all tasks using (CMSR). This paper also introduces a novel approach (4) that keeps the module architecture fixed, but evolves a separate routing for each task during training (CTR). Finally, approaches (2) and (4) are combined into (5), where both modules and task routing are coevolved (CMTR). Figure~\ref{fig:algs} gives high-level algorithmic descriptions of these methods, which are described in detail below.

\subsection{Coevolution of Modules}
\label{subsec:cm}

In Coevolution of Modules (CM), CoDeepNEAT is used to search for promising module architectures, which then are inserted into appropriate positions to create an \emph{enhanced soft ordering} network. The evolutionary process works as follows:
\begin{enumerate}
  \item \label{CM:zero} CoDeepNEAT initializes a population of modules $MP$. The blueprints are not used.
  \item Modules are randomly chosen from each species in $MP$, grouped into sets $M$ and are assembled into enhanced soft ordering networks.
  \item Each assembled network is trained/evaluated on some task and its performance is returned as fitness.
  \item Fitness is attributed to the modules, and NEAT evolutionary operators are applied to evolve the modules.
  \item The proess is repeated from step~\ref{CM:zero} until CoDeepNEAT terminates, i.e.\ no further progress is observed for a given number of generations.
\end{enumerate}
Unlike in soft ordering \cite{Meyerson:2018}, the number of modules and the depth of the network are not fixed but are evolved as global hyperparameters by CoDeepNEAT (however the layout is still a grid-like structure). Since the routing layout is fixed, the blueprint population of CoDeepNEAT, which determines how the modules are connected, is not used. Thus one key operation in the original CoDeepNEAT, i.e.\ inserting modules into each node of the blueprint DAG, is skipped; only the module population is evolved.

To assemble a network for fitness evaluation, an individual is randomly chosen from each species in the module population to form an ordered set of distinct modules $M$.
The hyperparameters evolved in each of the module's layers include the activation function, kernel size, number of filters, L2 regularization strength and output dropout rate.
In addition, CoDeepNEAT also coevolves global hyperparameters that are relevant to the entire assembled network as a whole; these include learning rate, the number of filters of the final layer of each module, and the weight initialization method. Evolvable hyperparameters in each node include the activation function, kernel size, number of filters, L2 regularization strength and output dropout rate. The modules are then transformed into actual neural networks by replacing each node in the DAG with the corresponding layer. To ensure compatibility between the inputs and outputs of each module, a linear $1 \times 1$ convolutional layer (number of filters determined by a global hyperparameter), followed by a max-pooling layer (provided that the feature map before pooling is at least $4 \times 4$) is included as the last layer in each module.

The modules are then inserted into the soft ordering network. The architecture of the network is interpreted as a grid of $K \times D$ slots, where $d$ indicates the depth of the network and the slots with the same $k$ value have the same module topology. For each available slot $T_{kd}$, the corresponding module $M_k$ is inserted. If $k > |M|$, then $M_{k \bmod |M|}$ is inserted instead.

Finally, each module in a particular slot has the potential to share its weights with modules that have the same architecture and are located in other slots of the blueprint. Flag $F_k$ in each module indicates whether or not the module's weights are shared. This flag is evolved as part of the module genotype in CoDeepNEAT. Also, there is also global flag $F_d$ for each depth of the soft ordering network. If the $M_k$ is placed in $T_{kd}$ and both $F_k$ and $F_d$ are turned on, then the module is able to share its weights with any other $M_k$ whose slot have both flags turned on as well. Such an arrangement allows each slot to have sharing enabled and disabled independently.

The assembled network is attached to separate encoders and decoders for each task and trained jointly using a gradient-based optimizer. Average performance over all tasks is returned as fitness back to CoDeepNEAT. That fitness is assigned to each of the modules in the assembled network. If a module is used in multiple assembled networks, their fitnesses are averaged into module fitness. After evaluation is complete, standard NEAT mutation, crossover, and speciation operators are applied to create the next generation of the module population \cite{stanley:ec02}.

\subsection{Coevolution of Modules/Shared Routing}
\label{subsec:cmsr}

Coevolution of Modules and Shared Routing (CMSR) extends CM to include blueprint evolution. Thus, the routing between various modules no longer follows the fixed grid-like structure, but instead an arbitrary DAG. Each node in the blueprint genotype points to a particular module species. During assembly, the blueprints are converted into deep multitask networks as follows:

\begin{enumerate}
  \item \label{CMSR:zero} For each blueprint in the population, an individual module is randomly chosen from each species.
  \item Each node in the blueprint is then replaced by the module from the appropriate species.
  \item If a module has multiple inputs from previous nodes in the blueprint, the inputs are soft merged first \cite{Meyerson:2018}.
	\item The process is repeated from step~\ref{CMSR:zero} until reaching a target number of assembled networks.
\end{enumerate}

As in CM, each node in the blueprint has a flag $F_i$ that indicates whether node $N_i$ should be shared or not. If two nodes are replaced by the same module and if both nodes have the sharing flag turned on, then the two modules will share weights. Such an arrangement allows each node to evolve independently whether to share weights or not. The training procedures for both CM and CMSR are otherwise identical. After fitness evaluation, the fitness is assigned to both blueprints and modules in the same manner as with CM. To accelerate evolution, the blueprint population is not initialized from minimally connected networks like the modules, but from randomly mutated networks that on average have five nodes.

\subsection{Coevolution of Task Routing}
\label{subsec:ctr}

This section introduces Coevolution of Task Routing (CTR), a multitask
architecture search approach that takes advantage of the dynamics of
soft ordering by evolving task-specific topologies instead of a single
blueprint.

\subsubsection{Overview}

Like in soft ordering, in CTR there are $K$ modules whose weights are shared everywhere they are used across all tasks.
Like in blueprint evolution, CTR searches for the best ways to assemble modules into complete networks.
However, unlike in blueprint evolution, CTR searches for a distinct module routing scheme \emph{for each task}, and trains a \emph{single set of modules} throughout evolution.
Having a distinct routing scheme for each task makes sense if the shared modules are seen as a set of building blocks that are assembled to meet the differing demands of different problems.
Training a single set of modules throughout evolution then makes sense as well: As modules are trained in different locations for different purposes during evolution, their functionality should become increasingly general, and it should thus become easier for them to adapt to the needs of a new location.
Such training is efficient since the core structure of the network need not be retrained from scratch at every generation.
In other words, CTR incurs no additional iterations of backpropagation over training a single fixed-topology multitask model.
Because of this feature, CTR is related to PathNet \cite{Fernando:2017}, which evolves pathways through modules as those modules are being trained.
However, unlike in PathNet, in CTR distinct routing schemes are coevolved across tasks, modules can be applied in any location, and module usage is adapted via soft merges.

CTR operates a variant of a $(1+1)$ evolutionary strategy ($(1+1)$-ES) for each task.
Separate ES for each task is possible because an evaluation of a multitask network yields a performance metric \emph{for each task}.
The $(1+1)$-ES is chosen because it is efficient and sufficiently powerful in experiments, though it can potentially be replaced by any population-based method.
To make it clear that a single set of modules is trained during evolution, and to disambiguate from the terminology of CoDeepNEAT, for CTR the term \emph{meta-iteration} is used in place of \emph{generation}.


\subsubsection{Algorithm Description}

Each individual constitutes a module routing scheme for a particular task.
At any point in evolution, the $i$th individual for the $t$th task is represented by a tuple $(\mathcal{E}_{ti}, G_{ti}, \mathcal{D}_{ti})$, where $\mathcal{E}_{ti}$ is an encoder, $G_{ti}$ is a DAG, which specifies the module routing scheme, and $\mathcal{D}_{ti}$ is a decoder.
The complete model for an individual is then given by
$$ \bm{y}_t = \big(\mathcal{D}_{ti} \circ \mathcal{R}\big(G_{ti}, \big\{\mathcal{M}_k\big\}_{k=1}^{K}\big) \circ \mathcal{E}_{ti}\big)(\bm{x}_t) \, ,$$
where $\mathcal{R}$ indicates the application of the shared modules  $\mathcal{M}_k$ based on the DAG $G_{ti}$.
Note that $\circ$ denotes function composition and $\mathcal{E}_{ti}$, and $\mathcal{D}_{ti}$ can be any neural network functions that are compatible with the set of shared modules.
In the experiments in this paper, each $\mathcal{E}_{ti}$ is an identity transformation layer, and each $\mathcal{D}_{ti}$ is a fully connected classification layer.

$G_{ti}$ is a DAG, whose single source node represents the input layer for that task, and whose single sink node represents the output layer, e.g., a classification layer.
All other nodes either point to a module $\mathcal{M}_k$ to be applied at that location, or a parameterless adapter layer that ensures adjacent modules are technically compatible.
In the experiments in this paper, all adapters are $2\times2$ max-pooling layers.
Whenever a node of $G_{ti}$ has multiple incoming edges, their contents are combined in a learned soft merge (Eq.~\ref{eq:soft_merge}).

The algorithm begins by initializing the shared modules $\{\mathcal{M}_k\}_{k=1}^{K}$ with random weights.
Then, each champion $(\mathcal{E}_{t1}, G_{t1}, \mathcal{D}_{t1})$ is initialized, with $\mathcal{E}_{t1}$ and $\mathcal{D}_{t1}$ initialized with random weights, and $G_{t1}$ according to some graph initialization policy.
For example, the initialization of $G_{t1}$ can be minimal or random.
In the experiments in this paper, $G_{t1}$ is initialized to reflect the classical deep multitask learning approach, i.e.,
$$\mathcal{E}_{t1} \rightarrow \mathcal{M}_1 \rightarrow \mathcal{M}_2 \rightarrow \ldots \rightarrow \mathcal{M}_K \rightarrow \mathcal{D}_{t1} \, ,$$
with adapters added as needed.

At the start of each meta-iteration, a challenger $(\mathcal{E}_{t2}, G_{t2}, \mathcal{D}_{t2})$ is generated by mutating the $t$th champion as follows (the insertion of adapters is omitted for clarity):
\begin{enumerate}
\item The challenger starts as a copy of the champion, including learned weights, i.e., $(\mathcal{E}_{t2}, G_{t2}, \mathcal{D}_{t2}) \coloneqq (\mathcal{E}_{t1}, G_{t1}, \mathcal{D}_{t1})$.
\item A pair of nodes $(u, v)$ is randomly selected from $G_{t2}$ such that $v$ is an ancestor of $u$.
\item A module $\mathcal{M}_k$ is randomly selected from $\{\mathcal{M}_k\}_{k=1}^{K}$.
\item A new node $w$ is added to $G_{t2}$ with $\mathcal{M}_k$ as its function.
\item New edges $(u, w)$ and $(w, v)$ are added to $G_{t2}$.
\item The scalar weight of $(w, v)$ is set such that its value after the softmax is some $\alpha \in (0, 1)$.
To initially preserve champion behavior, $\alpha$ is set to be small.
I.e., if $s_1, \ldots, s_m$ are the scales of the existing inbound edges to $v$, $s_{m+1}$ is the initial scale of the new edge, and $s_{\max} = \max(s_1, \ldots, s_m)$ then
$$ s_{m+1} = \ln\big(\frac{\alpha}{1-\alpha}\sum_{j=1..m} e^{s_j - s_{\max}}\big) + s_{\max} \, .$$

\end{enumerate}

After challengers are generated, all champions and challengers are trained jointly for $M$ iterations with a gradient-based optimizer.
Note that the scales of $G_{t1}$ and $G_{t2}$ diverge during training, as do the weights of $\mathcal{D}_{t1}$ and $\mathcal{D}_{t2}$. 
After training, all champions and challengers are evaluated on a validation set that is disjoint from the training data.
The fitness for each individual is its performance for its task on the validation set.
In this paper, accuracy is the performance metric. 
If the challenger has higher fitness than the champion, then the champion is replaced, i.e.,$(\mathcal{E}_{t1}, G_{t1}, \mathcal{D}_{t1}) \coloneqq (\mathcal{E}_{t2}, G_{t2}, \mathcal{D}_{t2})$.
After selection, if the average accuracy across all champions is the best achieved so far, the entire system is checkpointed, including the states of the modules.
After evolution, the champions and modules from the last checkpoint constitute the final trained model, and are evaluated on a held out test set.

\subsubsection{An Ecological Perspective}
\begin{figure}
  \begin{center}
    \includegraphics[width=\linewidth]{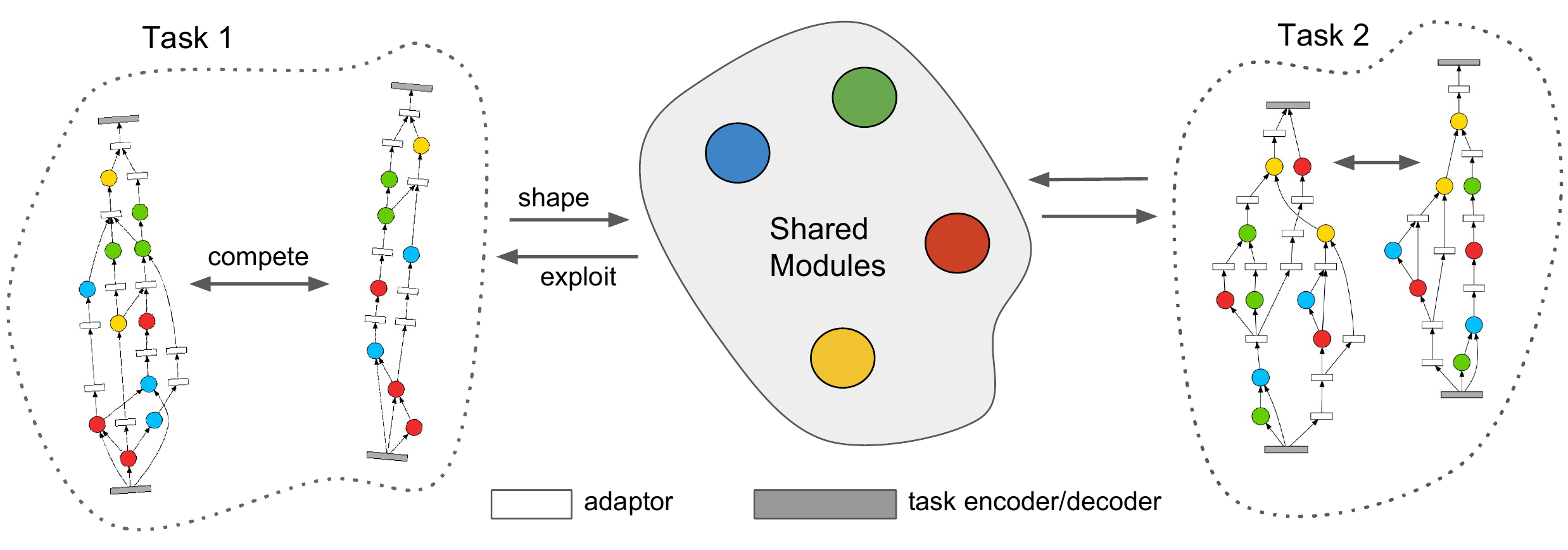}\\[-2ex]
    \caption{\label{fig:ctr_diagram} This figure shows an instance of CTR with three tasks and four modules that are shared across all tasks.
    Each individual assembles the modules in different ways.
    Through gradient-based training, individuals exploit the shared resources to compete within a task, and over time must develop mutualistic relationships with other tasks via their use of the shared modules.}
  \end{center}
\end{figure}

More than most evolutionary methods, this algorithm reflects an artificial ecology.
The shared modules can be viewed as a shared finite set of environmental resources that is constantly exploited and altered by the actions of different tasks, which can correspond to different species in an environment.
Within each task, individuals compete and cooperate to develop mutualistic relationships with the other tasks via their interaction with this shared environment.
A visualization of CTR under this perspective is shown in Figure~\ref{fig:ctr_diagram}.
Importantly, even if a challenger does not outperform its champion, its developmental (learning) process still affects the shared resources.
This perspective suggests a more optimistic view of evolution, in which individuals can have substantial positive effects on the future of the ecosystem even without reproducing.

\subsection{Coevolution of Modules and Task Routing}
\label{subsec:cmtr}

Both CM and CTR improve upon the performance of the original soft
ordering baseline. Interestingly, these improvements are largely
orthogonal, and they can be combined to form an even more powerful
algorithm called Coevolution of Modules and Task Routing (CMTR). Since
evolution in CTR occurs during training and is highly computational
efficient, it is feasible to use CoDeepNEAT as an outer evolutionary
loop to evolve modules. To evaluate and assign fitness to the modules,
they are passed on to CTR (the inner evolutionary loop) for evolving
and assembling the task specific routings. The performance of the
final task-specific routings is returned to CoDeepNEAT and attributed
to the modules in the same way as in CM: Each module is assigned the
mean of the fitnesses of all the CTR runs that made use of that
module.  Another way to characterize CMTR is that it overcomes the
weaknesses in both CM and CTR: CM's inability to create a customized
routing for each task and CTR's inability to search for better module
architectures.

CMTR's evolutionary loop works as follows:
\begin{enumerate}
  \item \label{CMTR:zero} CoDeepNEAT initializes a population of modules $MP$. The blueprints are not used.
  \item Modules are randomly chosen from each species in $MP$ and grouped together into sets of modules $M_k$.
  \item Each set of modules $M_k$ is given to CTR, which assembles the modules by evolving task-specific routings. The performance of the evolved routings on a task is returned as fitness.
  \item Fitness is attributed to the modules, and NEAT's evolutionary operators applied to evolve the modules.
  \item The process repeats from step~\ref{CMTR:zero} until CoDeepNEAT terminates, i.e. no improvement for a given number of generations.
\end{enumerate}
One difference between CMTR and CM is that each module's final convolutional layer has additional evolvable hyperparameters such as kernel size, activation function, and output dropout rate. Preliminary experiments suggested that the relatively complex routings in CMTR (when compared to CM and CMSR) require more complex final layers as well, thus evolving the complexity of the final layer is optimal. Like in CTR, the weights between modules are always shared in CMTR. If modules with completely new weights are added to the task routings, they have to be trained from scratch and may even hurt performance, whereas adding a module with already partially trained weights does not.
In addition, as the routings evolved by CTR are much larger than those discovered by CM and CMSR, disabling or evolving weight sharing significantly bloats the total number of weight parameters and slows training significantly.

\section{Experiments}
\label{sec:experiments}

This section details experiments comparing the five methods in the Omniglot MTL domain.

\subsection{Omniglot Character Recognition}

The Omniglot dataset consists of 50 alphabets of handwritten characters \cite{Lake:2015}, each of which induces its own character recognition task.
There are 20 instances of each character, each a $105 \times 105$ black and white image.
Omniglot is a good fit for MTL, because there is clear intuition that knowledge of several alphabets will make learning another one easier.
Omniglot has been used in an array of settings: generative modeling \cite{Lake:2015, Rezende:2016}, one-shot learning \cite{Koch:2015, Lake:2015, Shyam:2017}, and deep MTL \cite{Bilen:2017, Maclaurin:2015, Meyerson:2018, Rebuffi:2017, Yang:2017}.
Previous deep MTL approaches used random training/testing splits for
evaluation \cite{Bilen:2017, Meyerson:2018, Yang:2017}.  However, with
model search (i.e.\ when the model architecture is learned as well), a
validation set separate from the training and testing sets is needed.
Therefore, in the experiments in this paper, a fixed
training/validation/testing split of 50\%/20\%/30\% is introduced for
each task. Because training is slow and increases linearly with the
number of tasks, a subset of 20 tasks out of the 50 possible is used
in the current experiments. These tasks are trained in a fixed random
order.
Soft ordering is the current state-of-the-art method in this domain \cite{Meyerson:2018}. The experiments therefore use soft ordering as a starting point for designing further improvements.

\subsection{Experimental Setup}

For CoDeepNEAT fitness evaluations, all networks are trained using Adam \cite{Kingma:2014} for 3000 iterations over the 20 alphabets; for CTR, the network is trained for 120 meta-iterations (30,000 iterations). Each iteration is equivalent to one full forward and backward pass through the network with a single example image and label chosen randomly from each task.
The fitness assigned to each network is the average validation accuracy across the 20 tasks after training.

For CM and CMSR, CoDeepNEAT is initialized with approximately 50
modules (in four species) and 20 blueprints (in one species). For
CMTR, a smaller module population of around 25 (in two species) is
found to be beneficial in reducing noise since each module is evaluated more often.
During each generation, 100 networks are assembled from modules
and/or blueprints for evaluation. The global and layer-specific evolvable
hyperparameters are described in Section~\ref{sec:algorithms}.  With
CoDeepNEAT, the evaluation of assembled networks is distributed over
100 separate EC2 instances with a K80 GPU in AWS. The average time
for training is usually around 1-2 hours depending on the network
size. With CTR, because it is a $(1+1)$ evolutionary strategy with a
small population size, it is sufficient to run the algorithm on a
single GPU.

Because the fitness returned for each assembled network is noisy, to
find the best assembled CoDeepNEAT network, the top 50 highest fitness
networks from the entire history of the run are retrained for 30,000
iterations. For the CM and CMSR experiments, decaying the learning
rate by a factor of 10 after 10 and 20 epochs of training gave a
moderate boost to performance.  Similar boost is not observed for CTR
and CMTR and therefore learning rate is not decayed for them.  To
evaluate the performance of the best assembled network on the test set
(which is not seen during evolution or training), the network is
trained from scratch again for 30,000 iterations. For CTR and CMTR,
this is equivalent to training for 120 meta-iterations. During training, a
snapshot of the network is taken at the point of highest validation
accuracy. This snapshot is then evaluated and the average test
accuracy over all tasks returned.

\subsection{Results}

\begin{figure}[t]
  \begin{center}
    \includegraphics[width=\linewidth]{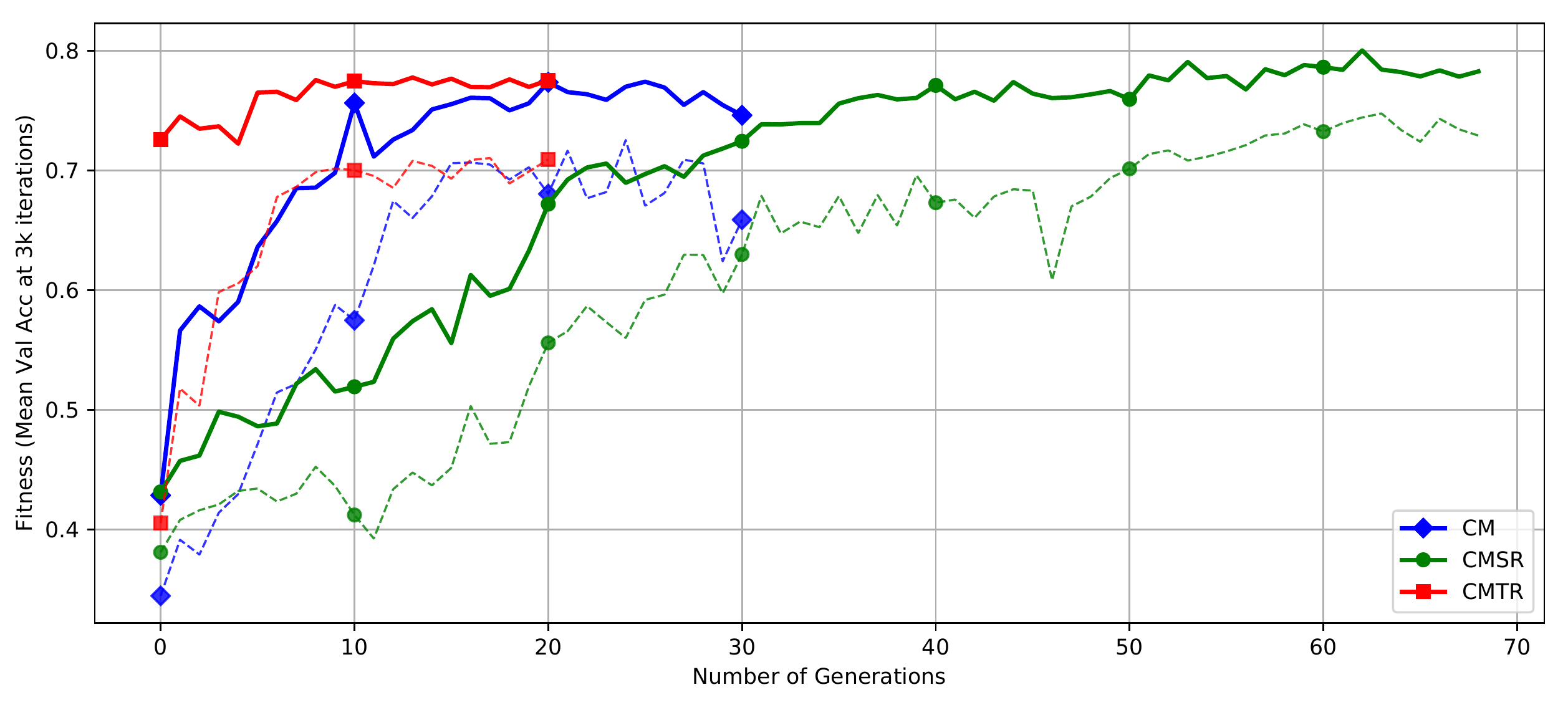}\\[-1.5ex]
    \vspace*{-2ex}
    \caption{Comparison of fitness (validation accuracy after partial training
      for 3000 iterations) over generations of single runs of CM, CMSR, and
      CMTR. Solid lines show the fitness of best assembled network and
      dotted line show the mean fitness. All methods reach a similar
      fitness, but CMTR is the fastest and CMSR the slowest.}
    \label{fg:evolution_plot}
  \end{center}
\end{figure}

\begin{figure}[t]
  \begin{center}
    \includegraphics[width=\linewidth]{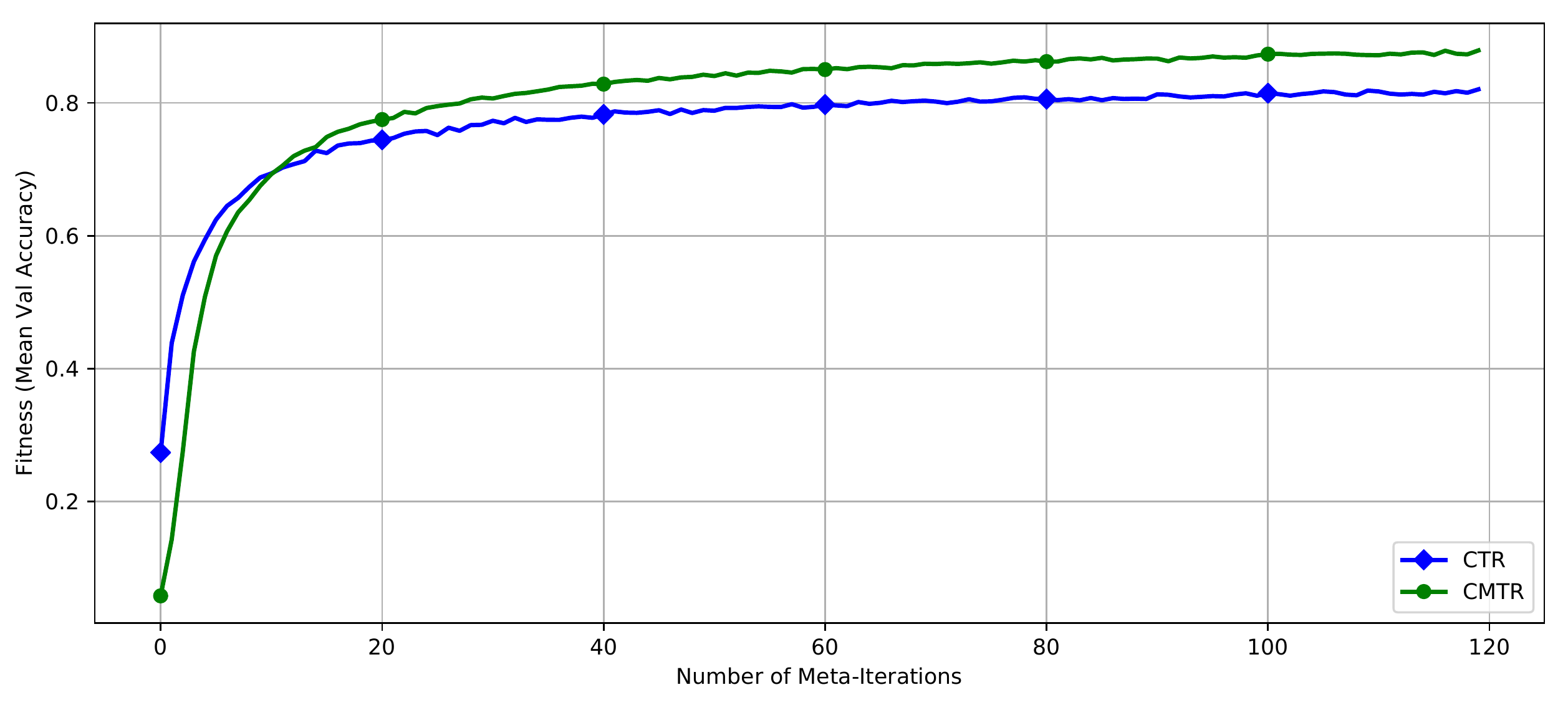}\\[-1.5ex]
    \vspace*{-2ex}
    \caption{Comparison of fitness over number of meta-iterations of
training for CTR and CMTR. Evolution discovers modules that leverage
the available training better, forming a synergy of the two processes.}
    \label{fg:evolution_ctr_plot}
  \end{center}
\end{figure}

\begin{figure}[t]
  \begin{center}
    \includegraphics[width=\linewidth]{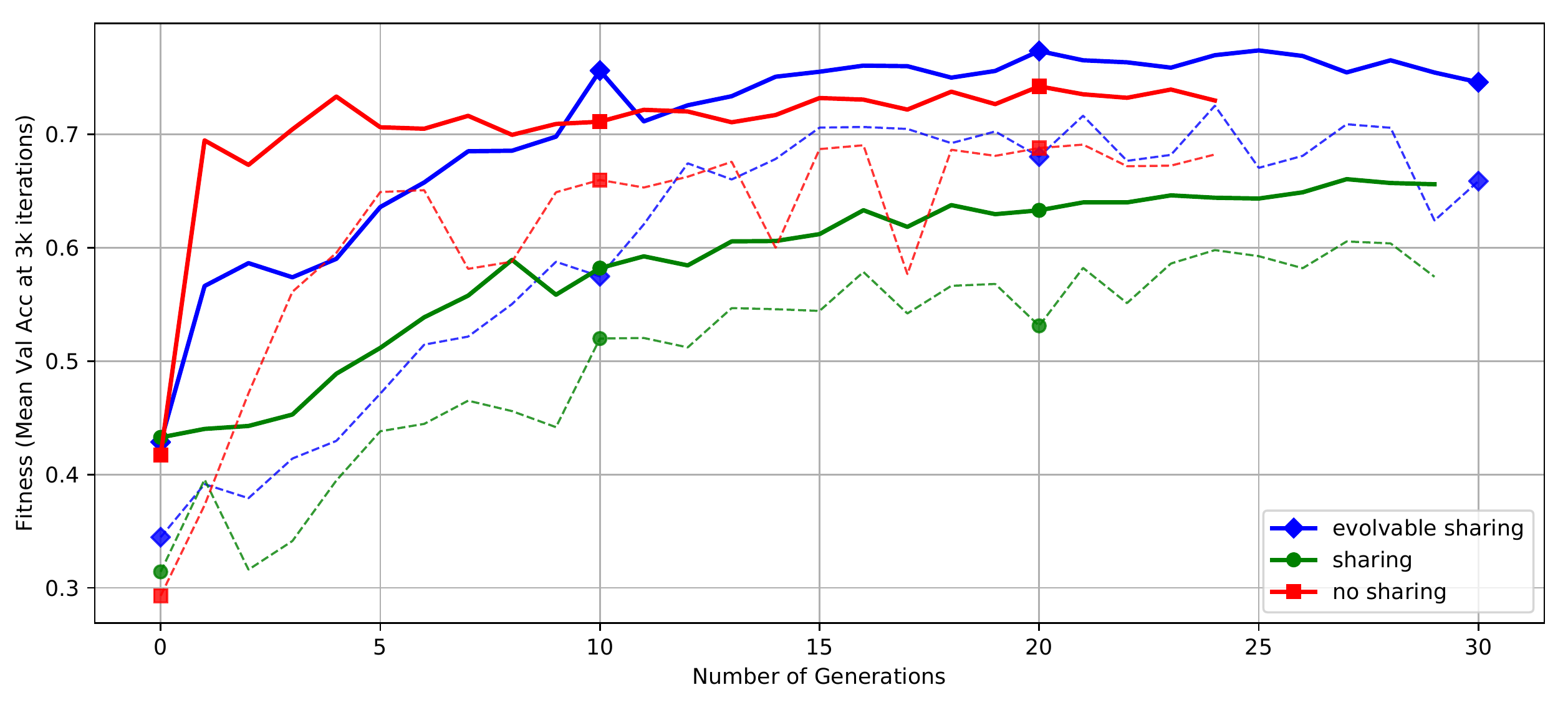}\\[-3ex]
    \caption{Comparison of fitness over generations of CM with disabling, enabling, and evolving module weight sharing. No sharing is better than forced sharing, but evolvable sharing outperforms them both, validating the approach.}
    \label{fg:evolution_plot_sharing}
  \end{center}
\end{figure}

\begin{table}[]
\small
  \centering
    \vspace*{3ex}
  \begin{tabular}{lcc}
    \toprule
    \textbf{Algorithm} & \textbf{Val Accuracy (\%)} & \textbf{Test Accuracy (\%)} \\
    \midrule
    1. Single Task \cite{Meyerson:2018}     & 63.59 (0.53) & 60.81 (0.50) \\
    2. Soft Ordering \cite{Meyerson:2018}   & 67.67 (0.74) & 66.59 (0.71) \\
    \midrule
    3. CM              & 80.38 (0.36)    & 81.33 (0.27) \\
    4. CMSR            & 83.69 (0.21)    & 83.82 (0.18) \\
    5. CTR             & 82.48 (0.21)    & 82.36 (0.19) \\
    6. CMTR            & \textbf{88.20} (1.02) & \textbf{87.82} (1.02) \\
    \bottomrule
  \end{tabular}
  \caption{Average validation and test accuracy over 20 tasks for each algorithm. CMTR performs the best as it combines both module and routing evolution. Pairwise \textit{t}-tests show all differences are statistically significant with $p<0.05$.}
  \label{fg:fitness}
\end{table}

Figure~\ref{fg:evolution_plot} demonstrates how the best and mean
fitness improves for CM, CMSR, and CMTR in the CoDeepNEAT outer loop
where module/blueprint coevolution occurs. All three algorithms
converge roughly to the same final fitness value, which is around 78\%
validation accuracy. CMTR converges the fastest, followed by CM, and
lastly CMSR. This result is expected since the search space of CMTR is
the smallest (only the modules are evolved with CoDeepNEAT), larger for CM (evolution
of modules and weight sharing), and largest for CMSR (evolution of
modules, blueprints, and weight sharing). Although CM, CMSR, and CMTR
converge to the same fitness in evolution, CMTR achieves better final
performance because training occurs via CTR.
Figure~\ref{fg:evolution_ctr_plot} compares how fitness (i.e.\ average
validation accuracy) improves for CTR (using the default modules) and
CMTR (using the best evolved modules discovered by CMTR) during
training, averaged over 10 runs. Interestingly, while CTR improves
faster in the first 10 meta-iterations, it is soon overtaken by CMTR,
demonstrating how evolution discovers modules that leverage the
available training better.

One open question is how much sharing of weights between
modules affects the performance of the assembled network. Although
disabling weight sharing is not optimal for CTR due to the complexity
of the routing, both CM and CMSR may benefit since their routing
topologies are much smaller (minimizing the effects of parameter
bloat). Figure~\ref{fg:evolution_plot_sharing} compares the effect of
enabling, disabling, and evolving weight sharing with CM.
Interestingly, disabling weight sharing leads to better performance
than enabling it, but evolving it is best. Thus, the design choice
of evolving sharing in CM and CMSR is vindicated. An analysis of the
architecture of the best assembled networks shows that weight sharing
in particular locations such as near the output decoders is a good
strategy.

Table~\ref{fg:fitness} shows the validation and test accuracy for the
best evolved network produced by each method, averaged over 10
runs. The best-performing methods are highlighted in bold and standard
error for the 10 runs is shown in parenthesis. In addition,
performance of the baseline methods are shown, namely (1) a
hand-designed single-task architecture, i.e.\ where each task is
trained and evaluated separately, and (2) the soft ordering network
architecture \cite{Meyerson:2018}. Indeed the methods improve upon the
baseline according to increasing complexity: Evolving modules and
evolving topologies is significantly better than the baselines, and
evolving both is significantly better than either alone. CMTR, the
combination of CoDeepNEAT and routing evolution, combines the
advantages of both and performs the best.

\begin{figure}[t]
  \begin{minipage}{0.45\linewidth}
    \centering
    \includegraphics[width=\linewidth]{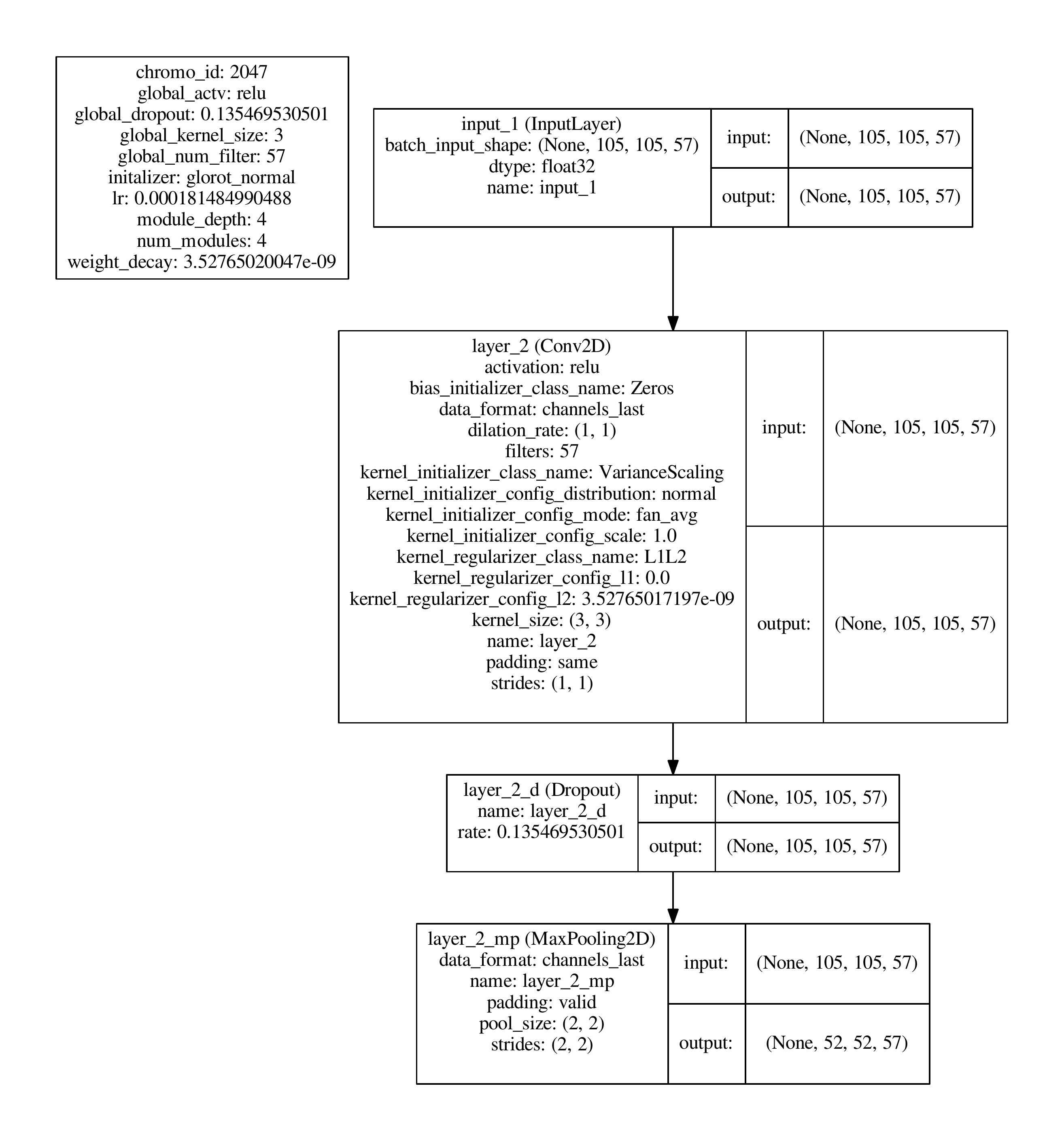}
    \subcaption{Topology of modules one and three.}
  \end{minipage}
  \begin{minipage}{0.45\linewidth}
    \centering
    \includegraphics[width=\linewidth]{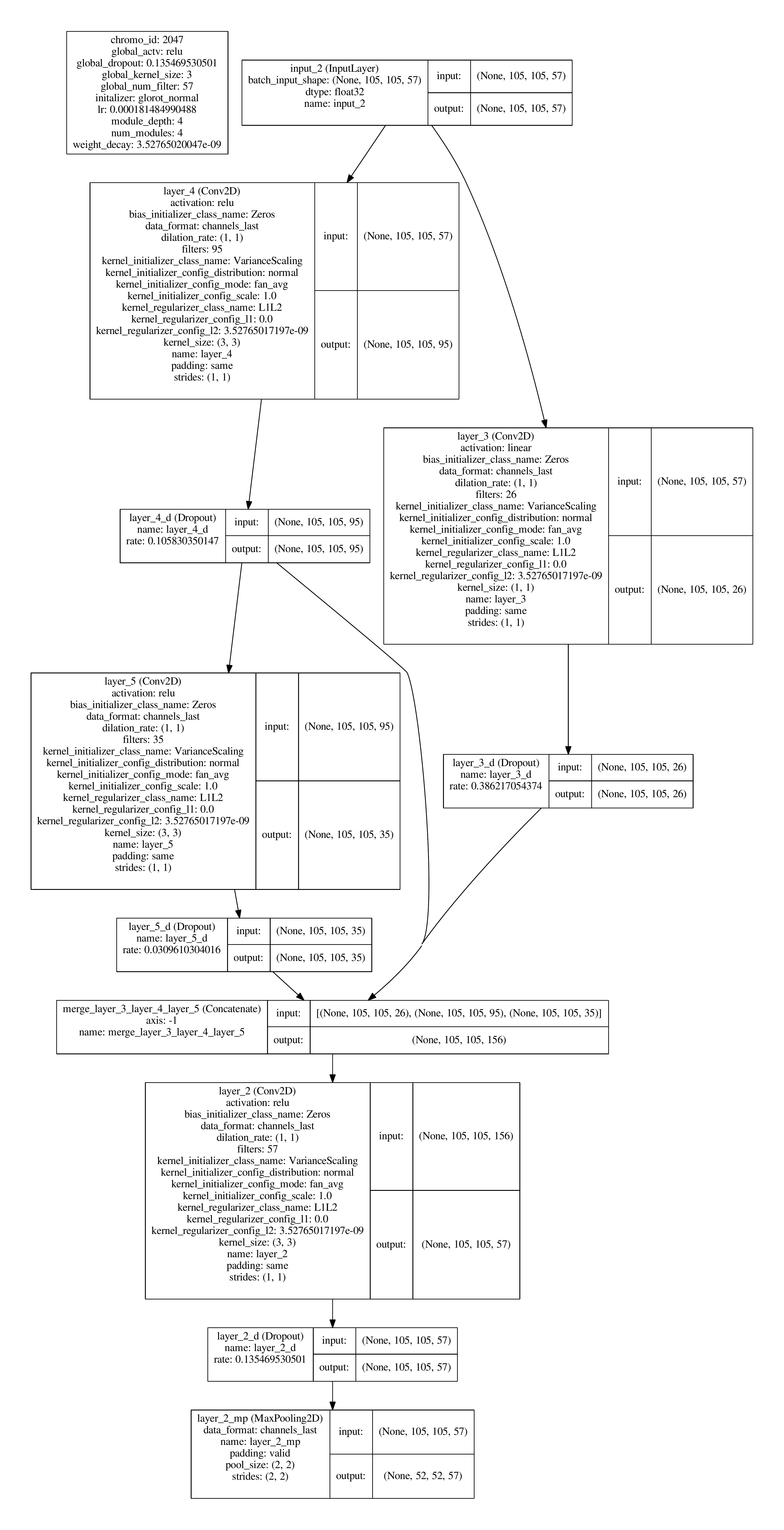}
    \subcaption{Topology of modules two and four.}
  \end{minipage}
  \begin{minipage}{0.45\linewidth}
    \centering
    \includegraphics[width=\linewidth]{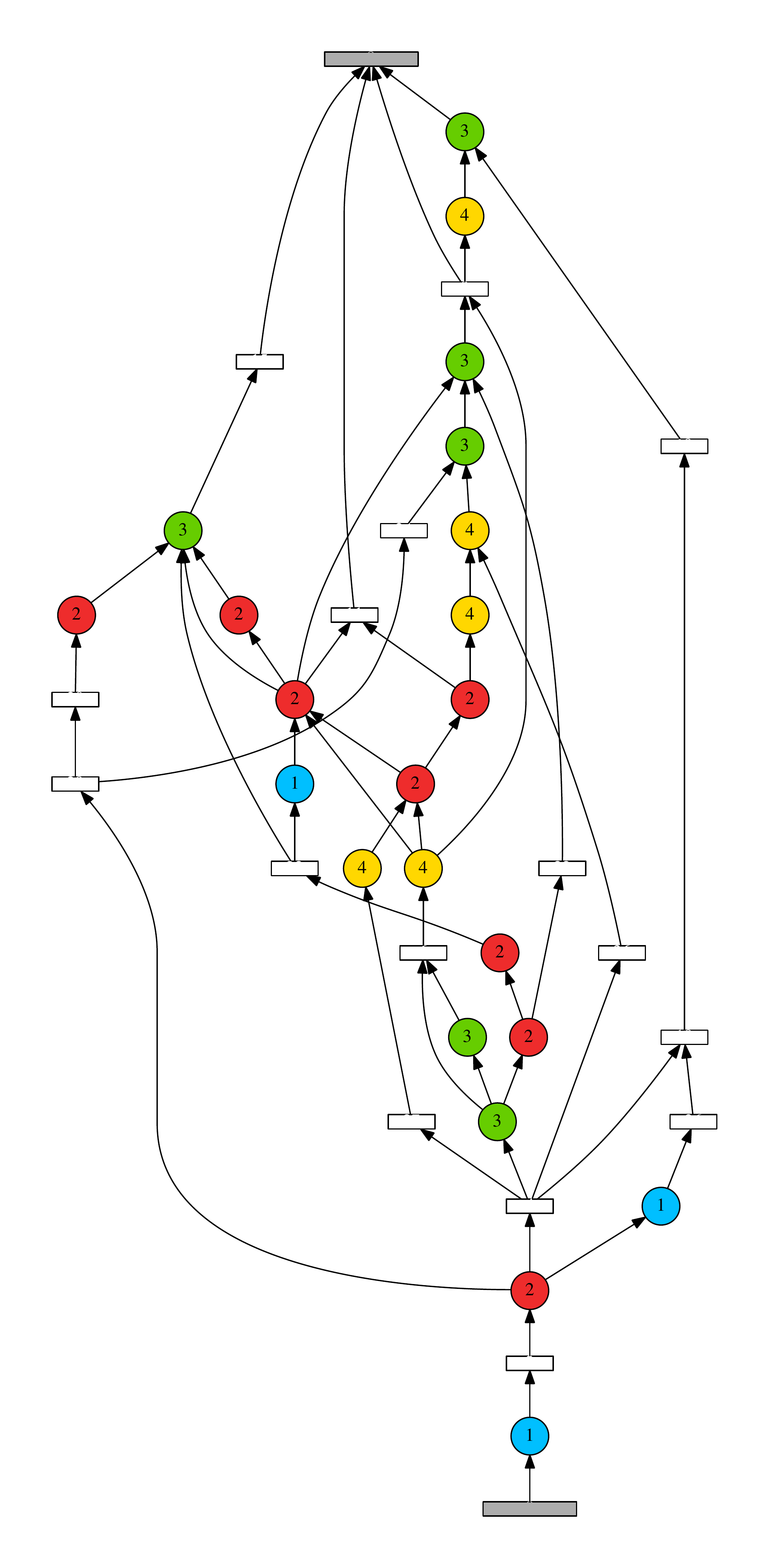}
    \subcaption{Routing for Angelic, an invented alphabet.}
  \end{minipage}
  \begin{minipage}{0.45\linewidth}
    \centering
    \includegraphics[width=0.45\linewidth]{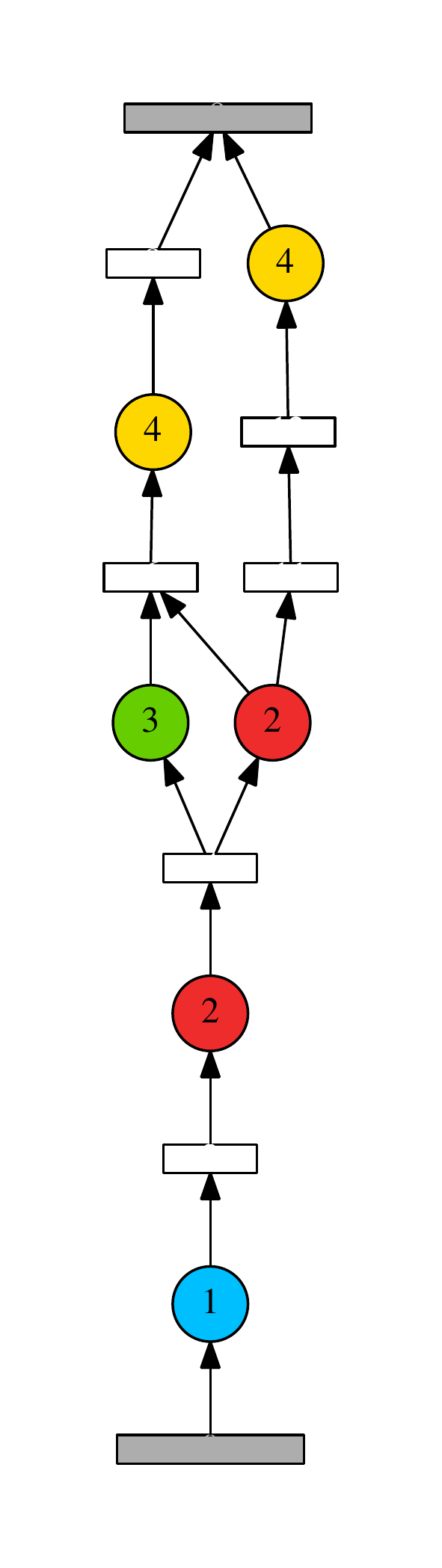}
    \includegraphics[width=0.45\linewidth]{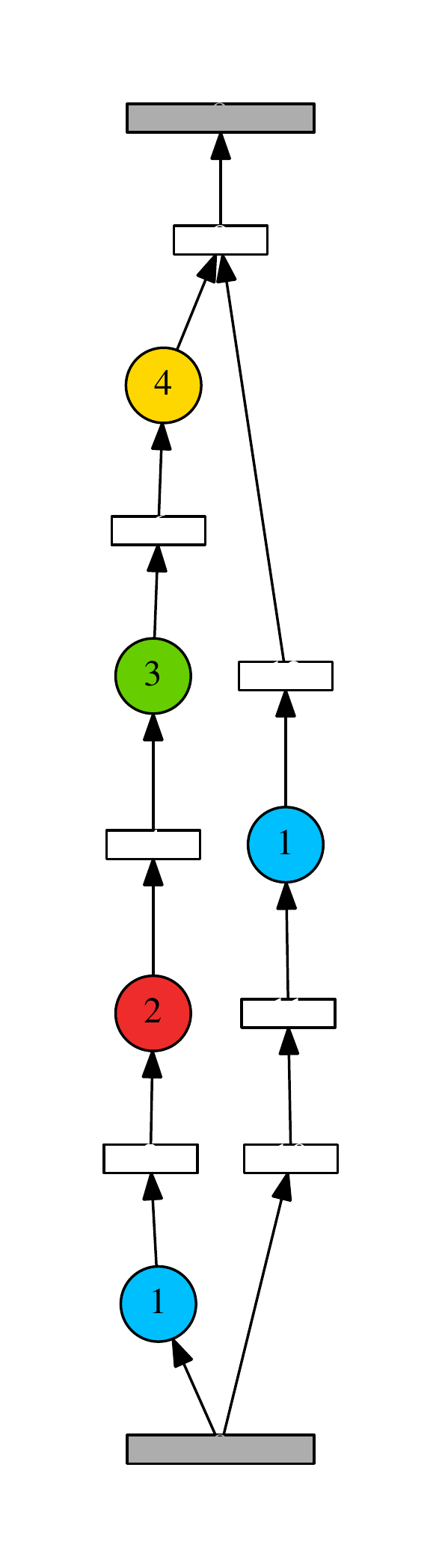}
    \subcaption{Routings for Gurmukhi and Mujarati.}
  \end{minipage}
  \caption{Structures of the best modules and routing topologies discovered by
    evolution. The two species in CMTR evolve very different
    modules: one simple and one complex. The thick boxes
    represent convolutional, medium max pooling, and thin
    dropout layers, with hyperparameters listed on the left. The
    routing topologies represent a range from simple to complex;
    similar alphabets have similar topologies, and the structure is
    consistently found. Such useful diversity would be difficult
    to develop by hand, demonstrating the power of evolution in
    designing complex systems.}
  \label{fg:vis}
\end{figure}

The best networks have approximately three million parameters.
Figure~\ref{fg:vis} visualizes one of the best performing modules from the
CMTR experiment, and sample routing topologies evolved for the
different alphabets. Because the CoDeepNEAT outer loop is based on two
species, the four modules passed to the CTR inner loop consist of two
different designs (but still separate weights). Thus, evolution
has discovered that a combination of simple and complex modules is
beneficial. Similarly, while the routing topologies for some
alphabets are simple, others are very complex. Moreover, similar
topologies emerge for similar alphabets (such as those that contain
prominent horizontal lines, like Gurmukhi and Manipuri). Also, when
evolution is run multiple times, similar topologies for the same
alphabet result. Such useful diversity in modules and routing
topologies, i.e.\ structures that complement each other and work well
together, would be remarkably difficult to develop by hand. However,
evolution discovers them consistently and effectively, demonstrating
the power of the approach.

\section{Discussion and Future Work}
\label{sec:discussion}

The experiments show that MTL can improve performance significantly across tasks, and that the architecture used for it matters a lot. Multiple ways of optimizing the architecture are proposed in this paper and the results lead to several insights.

First, modules used in the architecture can be optimized and the do end up different in a systematic way. Unlike in the original soft ordering architecture, evolution in CM, CMSR, and CMTR results in discovery of a wide variety of simple and complex modules, and they are often repeated in the architecture. Evolution thus discovers a useful set of building blocks that are diverse in structure.
Second, the routing of the modules matter as well. In CMSR, the shared but evolvable routing allows much more flexibility in how the modules can be reused and extends the principals that makes soft ordering useful. The power of CTR and CMTR is from evolving different topologies for different tasks, and tie the tasks together by sharing the modules in them. In addition, sharing components (including weight values) in CMTR is crucial to its performance. If indeed the power from multitasking comes from integrating requirements of multiple tasks, this integration will happen in the embeddings that the modules form, so it makes sense that sharing plays a central role.
Third, compared to the CTR and CMTR, CM and CMSR have evolved away from sharing of module weights, despite the fact that module architectures are often reused in the network. This result makes sense as well: because the topology is shared in this approach, the differentiation between tasks comes from differentiated modules. Such an approach is an opposite way to solve the problem. Even though it is an effective approach as well, it is not quite as powerful as differentiated topologies and shared modules.

There are several directions for future work. The proposed algorithms can be extended to many applications that lend themselves to the multitask approach. For instance, it will be interesting to see how it can be used to find synergies in different tasks in vision, and in language. Further, as has been shown in related work, the tasks do not even have to be closely related to gain the benefit from MTL. For instance, object recognition can be paired with caption generation. It is possible that the need to express the contents of an image in words will help object recognition, and vice versa.  Discovering ways to tie such multimodal tasks together should be a good opportunity for evolutionary optimization, and constitutes a most interesting direction for future work.

\section{Conclusion}
\label{sec:conclusion}

This paper presents a family of EAs for optimizing the architectures of deep multitask networks. They extend upon previous work which has shown that carefully designed routing and sharing of modules can significantly help multitask learning. The power of the proposed algorithms is shown by achieving state of the art performance on a widely used multitask learning dataset and benchmark.

\bibliographystyle{ACM-Reference-Format}
\bibliography{references}

\end{document}